\documentclass{article}
\usepackage{spconf,amsmath,graphicx}

\usepackage{arydshln}
\usepackage{todonotes}
\usepackage{url}
\usepackage{multirow}
\usepackage{amsfonts}
\usepackage{comment}
\usepackage{algorithm}
\usepackage[noend]{algpseudocode}

\title{PROTODA: EFFICIENT TRANSFER LEARNING FOR FEW-SHOT INTENT CLASSIFICATION}
\name{Manoj Kumar$^1$, Varun Kumar$^2$, Hadrien Glaude$^2$, Cyprien de Lichy$^2$, Aman Alok$^2$, Rahul Gupta$^2$}
\address{
  $^1$ Signal Analysis and Interpretation Lab, USC, Los Angeles, CA\\
  $^2$ Amazon Alexa, Cambridge, MA}

\begin{document}
%\ninept
%
\maketitle
\begin{abstract}
Practical sequence classification tasks in natural language processing often suffer from low training data availability for target classes. Recent works towards mitigating this problem have focused on transfer learning using embeddings pre-trained on often unrelated tasks, for instance, language modeling. 
We adopt an alternative approach by transfer learning on an ensemble of related tasks using prototypical networks under the meta-learning paradigm. 
Using intent classification as a case study, we demonstrate that increasing variability in training tasks can significantly improve classification performance.
Further, we apply data augmentation in conjunction with meta-learning to reduce sampling bias. 
We make use of a conditional generator for data augmentation that is trained directly using the meta-learning objective and simultaneously with prototypical networks, hence ensuring that data augmentation is customized to the task.
We explore augmentation in the sentence embedding space as well as prototypical embedding space. Combining meta-learning with augmentation provides upto 6.49\% and 8.53\% relative F1-score improvements over the best performing systems in the 5-shot and 10-shot learning, respectively.
\end{abstract}

\section{Introduction}
Intent classification (IC) is an important natural language processing task of voice controlled intelligent agents such as Amazon Alexa, Google Home, and Apple Siri. 
One of the first steps in such applications after converting speech to text is intent classification, where user queries are tagged with a sequence-level label identifying the underlying intent. 
To increase the capabilities of such agents, new intents are frequently added to the existing collection. Often, the development of a new intent starts with a few examples since labeled training data is scarce and expensive to obtain. 
Intent classification, in this case, resembles a few-shot learning setting where the goal is to generalize from a handful of training samples.

A natural resource available during new intent development is the collection of intents in-use by the voice controlled agent. These intents are often drawn from multiple domains such as music, reservations, dining, etc. and represent significant content variability between them.
While transfer learning from intents in-use tries to borrow high level feature representations during new intent development, it is prone to overfitting in the few-shot setting case.
On the contrary, learning from a diverse set of intents falls under the purview of meta-learning \cite{andrychowicz2016learning, ravi2017}, which learns across a collection of tasks as opposed to traditional supervised learning which learns across samples. Further, meta-learning has shown success in few-shot learning in computer vision \cite{ravi2017, finn2017model} and NLP \cite{yu2018diverse, GaoH0S19,huang-etal-2018-natural}, which can be useful for learning from a few annotated samples.

Another issue that is associated with few-shot learning is its susceptibility to sampling bias, i.e., estimated class distributions may not resemble the true population distribution due to limited sample availability.
A popular approach towards mitigating this bias is \emph{learning} to artificially synthesize new samples, known as data augmentation (DA).
DA has been an active research topic in NLP over the years.
A number of DA techniques have been experimented ranging from random perturbation~\cite{devries2017dataset} to deep-learning based approaches such as variability mode transfer between classes~\cite{hariharan2017low,schwartz2018delta}.
However, most of them perform augmentation independent of the task, i.e distribution of generated samples is independent of task loss. 

In this work, we optimize data augmentation model using task-specific objective, and combine it with meta learning to improve intent classification performance in the few-shot setting. 
In particular, we propose \emph{ProtoDA} which jointly trains a conditional generator network \cite{mirza2014conditional} with prototypical networks~\cite{snell2017prototypical,dwivediProtoGAN} (ProtoNets) to generate task specific samples. Combining data augmentation and ProtoNets is particularly suited, 
since during the few-shot setting prototypes are computed using a very limited number of examples, which incurs a sampling bias. Instead, computing prototype using both real and synthetic embedded examples allows to better estimate the class-specific population mean of the training set distribution for the class. 
We show the effectiveness of our method on intent classification task using open source datasets and a production scale corpora. 
Our primary contributions are: 1) combining meta-learning and data augmentation as an alternative to conventional transfer-learning specifically for low-resource IC, and 2) introducing data augmentation in the ProtoNet embedding space for improving task performance in NLP.
\section{Related work}

\subsection{Meta-learning using Prototypical Networks}

Meta-learning, or learning-to-learn, is a learning paradigm that learns at two-levels: within a task; and across multiple tasks while leveraging common knowledge among them.
The accumulated knowledge from an ensemble of tasks is used to improve few-shot accuracy on the target task, often on unseen classes. Various meta-learning approaches have been proposed mainly in the field of computer vision \cite{ravi2017,finn2017model}. 

ProtoNets ~\cite{snell2017prototypical} were first proposed for few-shot image classification with a relatively simple inductive bias when compared to other metric-learning methods such as matching networks \cite{vinyals2016matching} and relation networks \cite{sung2018learning}. ProtoNets are trained in an episodic manner using multiple tasks, with both tasks and train-test splits sampled within each episode (an episode refers to a single backpropagation step during the training process).
Few approaches exist which apply ProtoNets in NLP. In \cite{yu2018diverse}, ProtoNets were trained using a weighted sum of metrics to handle diverse tasks. In \cite{GaoH0S19}, the authors use an attention mechanism to weigh both features and samples during distance computation in the embedding space. In this work, we use the original formulation of ProtoNets and propose a joint data augmentation for the few-shot learning task.

%\textbf{I couldnt find the TL works from Amazon online - maybe they are not published yet?}

\subsection{Data Augmentation with Meta-Learning}

DA techniques in NLP have been explored on the lexical space including synonym replacement~\cite{zhang2015character}, back-translation~\cite{sennrich-etal-2016-improving} and sentence-level augmentation by replacing words with outputs from a language model~\cite{kobayashi2018contextual}.
Feature-space augmentation techniques on the other hand, generate new samples at the embedding space which are added to real samples during model training.
Applications for DA include natural language generation \cite{hu2017toward}, visual question answering \cite{kafle2017data}, relation classification \cite{xu2016improved} and machine translation \cite{fadaee2017data}. 
Most of the previous works focus on first training a data augmentation model, followed by adhoc data generation to augment the training set for the final task. Recently, \cite{wang2018low} proposed an end-to-end data \emph{hallucination} method that is trained using classification (task) loss. The hallucinator, a conditional generator, takes as inputs an original sample from the low-resource task and a noise sample, and generates a perturbed version of the samples which are augmented with the original training set. 
Inspired by this work, we explore a joint training of generator and classifier models in a meta-learning setup. 
The classifier (protonet in our case) is trained on the augmented set, including real as well as hallucinated samples. 
On the other hand, gradients from the classifier are propagated to the hallucinator for weight update. 
The feature extractor weights are typically pretrained and frozen while training the hallucinator. 
This setup encourages the hallucinator to generate samples that improve task performance and does not necessarily prioritize generating realistic-looking data samples.

\section{Modeling Setup}
Our modeling setup involves learning latent representations of text (e.g. sentence embeddings) using an encoder network, followed by classification using ProtoNets.
We consider adding the hallucinator at two separate points during training. 
We describe the setup of these components and the modeling architecture below. 

\subsection{Encoder Network}
\label{subsec:sent_enc}

Following~\cite{zhang2015character, chiu2016named, liang2017combining}, we use a neural network encoder in all our experiments to extract sentence-level embeddings from text. The encoder takes in a combination of character-level and word-level representations at the input. Character representations are learnt using a 2-D convolution neural network.
One-hot character encodings (of dimension 32) are passed through 2 convolutional layers (with a kernel size of 5) with max pooling followed by a temporal pooling layer to form a word-level representation. A dropout with a probability of 0.2 is used in both layers for regularization.
The CNN output is concatenated with pre-trained word-level embeddings (GloVe \cite{pennington2014glove}; 100-dimensional). The resulting word-level representation is then passed through a Bi-LSTM with a 128-dimensional hidden state to obtain contextualized word embeddings. A statistics-pooling layer computes the minimum, maximum and mean values of these embeddings to obtain sentence-level embeddings. 

\begin{figure}
%\centering
\includegraphics[width=0.48\textwidth, height=8cm]{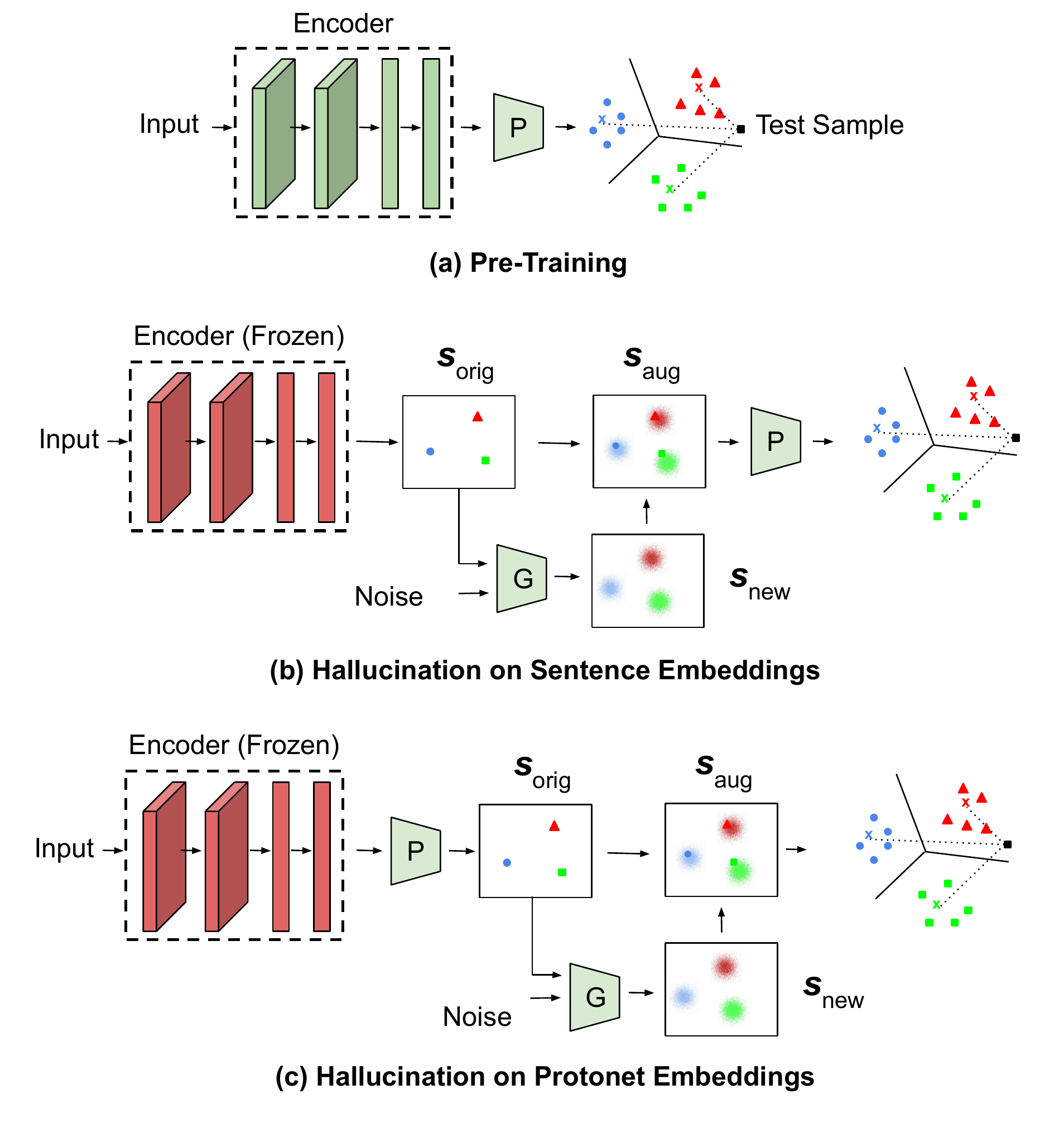}
\caption{Data augmentation in ProtoNets. The encoder is pre-trained using prototypical loss and it's weights frozen during hallucination. The hallucinator (G, a conditional generator) is trained with the ProtoNet's (P) loss function.}
\label{fig:aug_main}
\end{figure}
 
\subsection{Prototypical Networks}

ProtoNets learn a non-linear transformation where each class is reduced to a single point, specifically the centroid (prototype) of examples from that class. During inference a test sample is assigned to the class of nearest centroid.
%, similar to the nearest class mean method \cite{mensink2013distance}. 
Following, we illustrate a single episode of ProtoNet training, then extend it to multiple training tasks. 

\subsubsection{Episodic training}
\label{subsubsec:episodic_training}
Given a task $t$, consider a set of labeled training embeddings ${D_t = (\mathbf{X_{train}}, \mathbf{Y_{train}}) = {(\mathbf{x_1}, y_1), (\mathbf{x_2}, y_2), ...}}$\linebreak ${(\mathbf{{x_N}_{samples}}, {y_N}_{samples})}$ where $\mathbf{x_i} \in \mathbb{R}^M$ and $y_i \in \{1,2,..,C\}$. 
$D_t$ is sampled to form two sets: \textit{supports} ($S_t$) are used for prototype computation while \textit{queries} ($Q_t$) are used for estimating class posteriors and loss computation.
$S_t$ and $Q_t$ are not necessarily mutually exclusive.
ProtoNets learn a mapping $f_{\theta} : \mathbb{R}^M \rightarrow \mathbb{R}^P$ where the prototype of each class is computed as follows:
\begin{equation}
\label{eqn:proto_basic}
    v_{c} = \frac{1}{|S_{t,c}|} \sum_{(\mathbf{x_{i}},y_{i})\epsilon S_{t,c}} f_{\theta}(\mathbf{x_{i}})
\end{equation}
$S_{t,c}$ is the set of all examples in $S_t$ belonging to class $c$. For every test sample $\mathbf{x} \in Q_t$, the posterior probability given class $c$ is as follows:
\begin{equation}
\label{sofmax-eq}
    p(y=c | \mathbf{x})=\frac{\exp \left(-d\left(f_{\theta}(\mathbf{x}), \mathbf{v}_{c}\right)\right)}{\sum_{c^{\prime} \in C} \exp \left(-d\left(f_{\theta}(\mathbf{x}), \mathbf{v}_{c^{\prime}}\right)\right)}
\end{equation}
$d$ represents the distance function. %While the choice of $d$ can be arbitrary, 
Euclidean distance was chosen based on empirical results in the original ProtoNet implementation \cite{snell2017prototypical}. 
Learning proceeds by minimizing the negative log probability for the true class using gradient descent. Loss for the episode is computed as follows:
%\begin{equation}
%\label{eqn:proto_backprop}
% L(y,x) = -\;log(p_\theta(y=c \mid \mathbf{x})) 
%\end{equation}
\begin{equation}
\label{eqn:proto_backprop}
 L = - \frac{1}{|Q_t|} \sum_{(\mathbf{x_{i}},y_{i})\epsilon Q_t } \log(p_\theta(y_i=c \mid \mathbf{x_i})) 
\end{equation}

In general, a different task is chosen for each episode. ProtoNets, and meta-learning in general benefit from a large number of training tasks. 
%The original ProtoNet implementation \cite{snell2017prototypical} for instance, sampled 60 classes per task from a total of 4800 classes during training. 
In this work, we treat each training corpus as a task, and select all classes from the task within an episode. Pseudocode for ProtoNet training is provided in Algorithm \ref{algo:ProtoNet}.

%%%%%%%%%%%%%---------  ALGORITHM BLOCK --------- %%%%%%%%%%%%% 
\begin{algorithm}
\caption{Extending episodic learning to multiple tasks in the training corpus. SAMPLE$(S,K)$ denotes selecting $K$ samples uniformly at random from set $S$ with replacement.}
\label{algo:ProtoNet}
\hspace*{\algorithmicindent} \textbf{Input:} $T$: set of tasks, $N_{tasks}$: number of episodes
\begin{algorithmic}[1]
\For{i $\in \{1...N_{tasks}\}$}
    \State $t \gets $SAMPLE$(T,1)$. \Comment{Sample a task}
    \For{c in $\{1...C\}$}
        \State $D_{t,c} \gets$ Embeddings $\in$ class $c$ in task $t$
        \State $S_{t,c} \gets $SAMPLE$(D_{t,c},k)$ \Comment{k supports}
        \State $Q_{t,c} \gets $SAMPLE$(D_{t,c},q)$ \Comment{q queries}
    \EndFor
    \State Perform Episodic Training: Equations (\ref{eqn:proto_basic}-\ref{eqn:proto_backprop})
\EndFor
\end{algorithmic}
\end{algorithm}
%%%%%%%%%%%%%---------  ALGORITHM BLOCK --------- %%%%%%%%%%%%% 

The ProtoNet model architecture in this work consists of two feed-forward layers with 128 units in each layer. Hence, the model takes as input 768-dimensional embeddings from the sentence encoder and outputs 128-dimensional ProtoNet embeddings. Similar to the sentence encoder, dropout with a probability of 0.2 is used for regularization in both layers.

At each episode, we sample $k$ supports (the value of $k$ is experimented with 5 and 10) and 10 queries per class. The number of classes $C$ varies according to the task (i.e training corpus). The entire network is trained with Adam optimizer (lr=0.001, $\beta_1$=0.9, $\beta_2$=0.99) using the PyTorch toolkit.

\subsection{Hallucination}% (Sentence vs Prototype space)
\label{subsec:hallucination}

During meta-training, the learning setting (\textit{N-way, k-shot}: N classes, k samples/class) is carefully controlled to resemble the testing scenario. In \cite{snell2017prototypical} for instance, the authors matched the k-shot setting during both meta-training and meta-testing. While additional examples can be sampled from the class to reduce sampling bias during meta-training, data augmentation can introduce additional variations that are otherwise not present in the original data. 
In this work, we tie the augmentation process directly with the task objective (i.e intent classification). At every episode, gradients computed using the task loss are used to update not just the ProtoNet, but also the conditional generators used for data augmentation (hallucination).

Let $S_{c,orig}$ represent supports from class $c$ during an episode of meta-training. Let $S_{c,new}$ represent a subset of $S_{c,orig}$ chosen for augmentation. Each sample $\in S_{c,new}$ is passed as input to a generator network ($G$) alongwith a noise vector to produce an augmented sample. 
%Let $S_{c,new}$ represent the generated set of examples. 
The new prototype for class $c$ is computed as centroid of $S_{c,aug} = S_{c,orig} \cup S_{c,new}$:
\begin{equation}
    v_{c,aug} = \frac{1}{|S_{c,aug}|} \left( \sum_{\mathbf{e}_i \in S_{c,orig}} \mkern-18mu f_{\theta}(\mathbf{e}_i) + \sum_{\mathbf{e}_j \in S_{c,new}} \mkern-18mu f_{\theta}(G(\mathbf{e}_j, \mathbf{z})) \right)
\end{equation}
where $\mathbf{e} = E(\mathbf{x})$, $E$ represents the sentence encoder described in Section \ref{subsec:sent_enc} and $\mathbf{z}$ is a noise vector with same dimensionality as $\mathbf{e}$. 
%Note that 
The generator training does not have a separate objective of its own, but updates according to the episodic loss in Equation (\ref{eqn:proto_backprop}). While the method described above augments samples in the sentence embedding space (Figure \ref{fig:aug_main}b), an alternative approach is to augment samples in the ProtoNet embedding space (Figure \ref{fig:aug_main}c), i.e
\begin{equation}
    v_{c,aug} = \frac{1}{|S_{c,aug}|} \left( \sum_{\mathbf{e}_i \in S_{c,orig}} \mkern-18mu f_{\theta}(\mathbf{e}_i) + \sum_{\mathbf{e}_j \in S_{c,new}} \mkern-18mu G(f_{\theta}(\mathbf{e}_j), \mathbf{z}) \right)
\end{equation}

We meta-train the hallucinator network as follows: First, the sentence encoder and ProtoNet are pre-trained for 20000 episodes. Next, the sentence encoder weights are frozen, while the generator network (two feed-forward layers with 128 units in each layer, dropout with a probability of 0.2) and ProtoNet are trained together for another 20000 episodes. Following \cite{wang2018low}, the hallucinator weights are initialized with block diagonal identity matrices. At each episode, 20\% of the original samples are randomly selected for hallucination, hence $|S_{c,aug}| = 1.2\times|S_{c,orig}|$. The generator weights are frozen during meta-testing, and used to augment the supports for prototype computation.
\section{Datasets}
\label{sec:datasets}

We use two source corpora: the task-oriented dialog corpus from Facebook (FB) \cite{gupta2018semanticparsing} containing crowd-sourced annotations for queries from the navigation and event management domains, and
the Air Travel Information System (ATIS) corpus \cite{hemphill1990atis} consisting of spoken queries from the air travel domain. Both corpora contain natural language queries (as opposed to written form) and are more suitable to our target domain, i.e voice-controlled agents.
We remove intents with less than 20 utterances from both corpora and utterances with multiple root intents from FB. This results in a total of 45,489 utterances from 25 intents. 

For evaluation purpose, we use the SNIPS corpus \cite{coucke2018snips} which has served as a benchmark for recent sequence classification tasks in NLP. SNIPS contains crowd-sourced queries from seven intents with a balanced sample distribution across classes: $\approx$ 2000 samples for training and 100 samples for validation for each intent. 
We divide the seven intents in SNIPS into train (BookRestaurant, AddToPlaylist, RateBook, SearchScreeningEvent) and test (PlayMusic, GetWeather, SearchCreativeWork) to evaluate within different experimental configurations (see Section \ref{subsec:exp1})

Similar to previous meta-learning works in computer vision  \cite{ravi2017,snell2017prototypical} which have used hundreds of classes during meta-training, we curate an Alexa corpus to aid the unseen intent configuration (Section \ref{subsec:exp1}).
The Alexa corpus consists of user queries directed at the devices supported by the smart agent. Voice queries were manually transcribed and labeled for intents. The queries span 68 Alexa third-party \emph{skills}\footnote{https://developer.amazon.com/en-US/alexa/alexa-skills-kit} and $\approx$1100 intents. The number of intents per skill ranges between $2$ to $30$. Treating each Alexa skill as a task (Section \ref{subsubsec:episodic_training}), the augmented training corpus containing FB, ATIS, SNIPS and Alexa contains $71$ training tasks. 

\begin{table}[h]
\centering
\caption{Training intents used in each experimental setup. SNIPS-4 refers to BookRestaurant, AddToPlaylist, RateBook, SearchScreeningEvent classes.  }
\label{tab:exp1_plan}
%\resizebox{0.45\textwidth}{!}{%
\begin{tabular}{ccc} \hline
              & Seen Intents          & Unseen Intents \\ \hline
\rule{0pt}{2ex}           
Single Task & SNIPS (All)           & SNIPS-4  \\ [1mm]
Multi Task & \begin{tabular}[c]{@{}c@{}}FB,ATIS\\ + SNIPS (All)\end{tabular} & \begin{tabular}[c]{@{}c@{}}FB,ATIS\\ + SNIPS-4\end{tabular} \\ \hline 
\end{tabular}
\end{table}

\setcounter{table}{3}
\begin{table*}[t]
\centering
\caption{Micro-F1 scores (\%) using different augmentation methods (None, Noise: standard normal, Hall: Hallucination) and embeddings (Sent: Sentence, Proto: Protonet) for transfer learning on seen and unseen intents. $\pm$ indicates 95\% confidence interval.}
\vspace{-3mm}
\label{tab:aug_results}
\resizebox{\textwidth}{!}{%
\begin{tabular}{l:cccc:cccc} \hline
 & \multicolumn{2}{c}{Seen (5-shot)} & \multicolumn{2}{c}{Seen (10-shot)} & \multicolumn{2}{c}{Unseen (5-shot)} & \multicolumn{2}{c}{Unseen (10-shot)} \\ 
Augmentation & Sent & Proto & Sent & Proto & Sent & Proto & Sent & Proto \\ [1pt] \hline
None & \multicolumn{2}{c}{75.60 $\pm$ 4.27} & \multicolumn{2}{c}{86.40 $\pm$ 1.91} & \multicolumn{2}{c}{79.85 $\pm$ 1.43} & \multicolumn{2}{c}{89.02 $\pm$ 1.24} \\
Noise & 75.72 $\pm$ 3.63 & \textbf{77.47 $\pm$ 3.66} & 86.85 $\pm$ 1.95 & 86.93 $\pm$ 1.93 & 80.57 $\pm$ 1.74 & 82.17 $\pm$ 1.42 & 89.87  $\pm$ 1.00 & 89.93 $\pm$ 0.79 \\
Hall & \textbf{76.30 $\pm$ 3.10} & 76.62 $\pm$ 3.52 & \textbf{87.11 $\pm$ 1.88} & \textbf{ 88.08 $\pm$ 1.88} & \textbf{81.33 $\pm$ 1.87} & \textbf{83.67 $\pm$ 1.65} & \textbf{90.38 $\pm$ 0.87} & \textbf{91.18 $\pm$ 0.73} \\ \hline
\end{tabular}}
\end{table*}

\section{Experiments}

\subsection{Protonets for Transfer Learning}
\label{subsec:exp1}

In the first set of experiments, we evaluate protonets for transfer learning under two conditions: seen intents and unseen intents.
For seen intents, we make use of the train partitions from test intents for model backpropagation.
This may not be always possible, when for instance, on-device computation for model adaptation is restricted/infeasible. Nevertheless, these experiments analyze the value of including related tasks during ProtoNet training. 
We repeat the experiments by removing test intents during training time, which more accurately represents the scenario where we wish to evaluate pre-trained models on newly-introduced skills (intents) for a voice-controlled assistant. 
For the case of seen intents, we develop a competitive conventional transfer learning method (Conv TL) to compare with ProtoNets - 
We use the sentence encoder described in Section \ref{subsec:sent_enc} and add two feed-forward layers (128 units in each layer, dropout with probability of 0.2) similar to the ProtoNet architecture. The model is trained to minimize cross-entropy loss on FB, ATIS and SNIPS (train intents). 
Following, the training partition from the test intents are used to fine-tune the network by replacing the final softmax layer.

For both seen and unseen intents, we experiment with two different setups by varying the number of tasks available during meta-training. Under the single-task setup, we use only the SNIPS corpus during meta-training. Here, transfer learning happens in the case of unseen intents. 
Under the multi-task setup we make use of 3 tasks - FB, ATIS and SNIPS corpora. Increasing the number of tasks is expected to improve the learning capability of ProtoNets.
Table \ref{tab:exp1_plan} illustrates the proposed experiment setups.

\subsection{Data Augmentation}
\label{subsec:exp2}
% Write about data augmentation

In the next set of experiments, we select the best performing configurations from seen and unseen intents cases and perform data hallucination (Hall) during training. 
As mentioned in Section \ref{subsec:hallucination}, we train the hallucinator at one of two spaces, sentence embeddings or ProtoNet embeddings. 
At each space, we compare hallucination with random perturbation data augmentation (Noise) which has been shown as a competitive baseline \cite{kumar2019closer} in few-shot data augmentation experiments. 
Moreover, we control the amount of random perturbation per class to match that of hallucination (20\% i.e., we introduce one synthetic embedding for every 5 real embeddings) thereby creating a fair comparison with hallucination.
In this method, we augment the embeddings with additive and multiplicative noise generated using a normal distribution with zero-mean and standard variance of 10\% batch variance in every dimension.

During evaluation, all experiments including the baseline are repeated for 20 trials by randomly selecting $k$ (= 5,10) labeled examples from the train partitions for prototype computation. The validation partitions from test intents are used for evaluation. We report the averaged micro F1 scores alongwith the 95\% confidence intervals for all experiments.

\setcounter{table}{1}
\begin{table}[h]
\centering
\caption{Micro-F1 scores (\%) for TL with seen intents}\smallskip
\label{tab:results_seenIntents}
%\resizebox{0.48\textwidth}{!}{%
\begin{tabular}{ccc} \hline
Method & \multicolumn{1}{c}{5-shot} & \multicolumn{1}{c}{10-shot} \\ \hline
\rule{0pt}{2ex} Conv TL & 74.98 $\pm$ 3.46  & 82.02 $\pm$ 3.94 \\
Single Task & 70.48 $\pm$ 4.20 & 82.48 $\pm$ 3.27 \\
Multi Task & \textbf{75.60} $\pm$ \textbf{4.27} & \textbf{86.40} $\pm$ \textbf{1.91} \\ \hline
\end{tabular}
\end{table}

\setcounter{table}{2}
\begin{table}[h]
\centering
\caption{Micro-F1 scores (\%) for TL with unseen intents}\smallskip
\label{tab:results_unseenIntents}
%\resizebox{0.48\textwidth}{!}{%
\begin{tabular}{ccc} \hline
Method & \multicolumn{1}{c}{5-shot} & \multicolumn{1}{c}{10-shot} \\ \hline
%Conv TL & 74.98 $\pm$ 3.46  & 82.02 $\pm$ 3.94 \\
\rule{0pt}{2ex} Single Task & 49.95 $\pm$ 4.79 & 57.45 $\pm$ 3.33 \\
Multi Task & 70.82 $\pm$ 0.97 & 73.18 $\pm$ 0.54 \\ 
%\begin{tabular}[c]{@{}c@{}}Across-Domain\\ + Alexa\end{tabular} & \textbf{79.85 $\pm$ 1.43} & \textbf{89.02 $\pm$ 1.24} \\ \hline
Across-Domain + Alexa & \textbf{79.85 $\pm$ 1.43} & \textbf{89.02 $\pm$ 1.24} \\ \hline
\end{tabular}
\end{table}

\section{Results and Discussion}

Tables \ref{tab:results_seenIntents} and \ref{tab:results_unseenIntents} show the performance on seen intents and unseen intents respectively. 
In the former, we observe that single task transfer does not provide significant gains over ConvTL, even failing to outperform in the 5-shot case. 
We observe that ProtoNets benefit with increased task variability during multiple tasks, where transfer learning happens across corpora. 
While unseen intents in general result in lesser classification performance owing to non-availability of classes during training, gains from increased task variability (during multiple tasks) are significant. Specifically, the 5-shot and 10-shot settings result in 20.87\% and 15.73\% absolute improvement over the single-task, in comparison to 5.12\% and 3.92\% during seen intents.
When the number of training tasks is greatly increased using the Alexa corpus, the gains in classification outperform the best performing models in seen intents, including ConvTL. These results demonstrate the importance of variability in training tasks for meta-learning.

In Table \ref{tab:aug_results}, we see that both augmentation methods improve performance across the different settings. In most cases, hallucination improves over random perturbations, as it can learn to de-bias prototypes computed from a very small set of examples. Within each TL method and $k$-shot setting, ProtoNet embeddings prove to be a better choice for DA over sentence embeddings. We believe that the smaller dimensionality in the ProtoNet space and proximity to the training objective function (ProtoNet loss) makes hallucination more effective in the protonet embedding space.

\section{Conclusion}

Conventional TL approaches for low-resource NLU applications are still dependent on a small number of labeled samples from unseen intents during model training. In this work, we propose an alternative approach by combining meta-learning with data hallucination. Given sufficient variability in the training set (represented as tasks), we show that ProtoNets outperform models trained with standard cross-entropy objectives. Data augmentation further assists generalization by reducing sampling bias during prototype computation.
While augmenting samples with additive and multiplicative noise is beneficial, we show better improvements by learning to optimize the hallucinator directly with the task loss.
In the future, we would like to extend this approach to downstream NLU tasks for voice controlled agents, such as named entity recognition which entails sequence labels. 

\bibliographystyle{IEEEbib}
\bibliography{protoda}

\end{document}